\documentclass{article}
\usepackage{spconf,amsmath,graphicx}
\usepackage[figureposition=bottom,tableposition=top]{caption}
\usepackage{multicol}
\usepackage{booktabs}
\usepackage{xcolor}

\title{Scale-Invariant Multi-Oriented Text Detection in Wild Scene Images }

\name{Kinjal Dasgupta\textsuperscript{1}, Sudip Das\textsuperscript{2}\sthanks{Correspondace Author}, Ujjwal Bhattacharya\textsuperscript{2}}

\address{\textsuperscript{1}Heritage Institute of Technology, Kolkata \hspace{0.1cm}
\textsuperscript{2}Indian Statistical Institute, Kolkata \hspace{0.1cm}
\\
{\tt\small kinjal.dasgupta.9@gmail.com,  d.sudip47@gmail.com, ujjwal@isical.ac.in}
}

\begin{document}
\maketitle
\begin{abstract}

\name{Author(s) Name(s)}
\address{Author Affiliation(s)}
\sthanks{}
Automatic detection of scene texts in the wild is a challenging problem, particularly due to the difficulties in handling (i) occlusions of varying percentages, (ii) widely different scales and orientations, (iii) severe degradations in the image quality etc. In this article, we propose a fully convolutional neural network architecture consisting of a novel \textit{Feature Representation Block} (FRB) capable of efficient abstraction of information. The proposed network has been trained using curriculum learning with respect to difficulties in image samples and gradual pixel-wise blurring. It is capable of detecting texts of different scales and orientations suffered by blurring from multiple possible sources, non-uniform illumination as well as partial occlusions of varying percentages. Text detection performance of the proposed framework on various benchmark sample databases including \textit{ICDAR 2015}, \textit{ICDAR 2017 MLT}, \textit{COCO-Text} and \textit{MSRA-TD500} improves respective state-of-the-art results significantly.  Source code of the proposed architecture will be made available at \textcolor{red}{github}.

\end{abstract}

\section{Introduction}

\begin{figure}[t]
 \includegraphics[height=1.1in,width=1.8in]{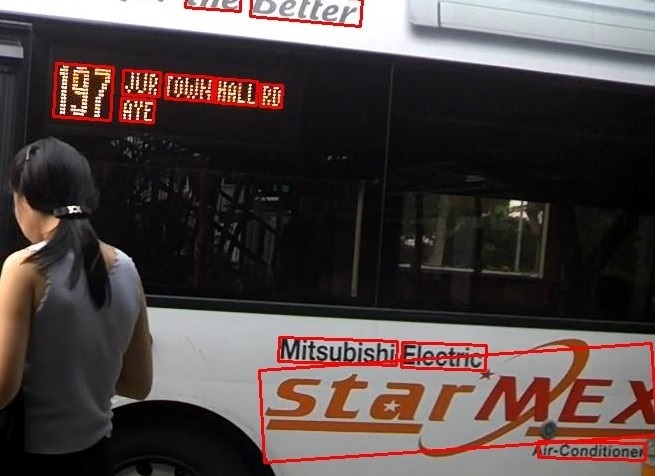}
 \includegraphics[height=1.1in,width=1.54in]{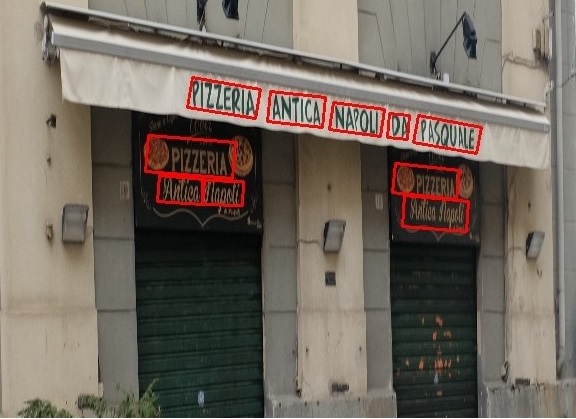}
\caption{Red rectangular boxes show widely varying scene texts detected by the proposed framework.} \label{fig:2samples_of_ICDAR_MLT}
\end{figure}

Although existing state-of-the-art scene text detection models have already acquired promising performances on moderately well-behaved scene image samples, till date, efficient detection of incidental scene texts (such as texts captured by wearable cameras where the capture is difficult to control) remains one of the most challenging tasks in the Computer Vision community.  In 2015, the ICDAR Robust Reading Competition \cite{karatzas2015icdar} introduced a few challenges of processing incidental scene texts and published a related sample database for the first time. COCO-Text \cite{veit2016coco}, another benchmark dataset created at a later period also contains similar samples of complex everyday scenes. Such samples are often motion blurred, non-uniformly illuminated, partially occluded, multi-oriented or multi-scaled. Efficient detection of similar texts in scene image needs further intensive studies. Robustness of the detection procedure largely depends on the distinguishing power of the feature representation between text and non-text components. Since supremacy of deep learning-based strategies for feature representation of raw images over various traditional hand-crafted filters has already been established \cite{luan2018gabor}, our present study is centred around the development of an efficient deep convolutional neural network architecture for the present purpose. The deep architecture developed in this study includes a novel \textit{Feature Representation Block} (FRB) for robust detection of scene texts. This FRB has been suitably designed towards abstraction of information at multiple levels generating features adaptive to changes in orientation and scale. Its Gabor Filter based Convolutional component captures scale and orientation information, the channel-wise attention map enhances important features and attenuates noisy and unwanted background information, the 4Dir IRNN component \cite{Bell_2016_CVPR} takes care of contextual information by moving the RNNs laterally across the image while the \textit{conditional random fields} (CRFs) \cite{Liu_2019_ICCV} based aggregation component refines inbound information of multiple scales. Figure. \ref{fig:2samples_of_ICDAR_MLT} shows outputs of the proposed detection framework on two samples of Robust Reading Competition datasets \cite{karatzas2015icdar,nayef2017icdar2017} containing incidental scene texts.


\section{Related Works}
Detection and recognition of texts in scene images have been studied for a long period. A recent survey of existing approaches and analysis of results can be found in \cite{zhu2016survey}. Among the traditional approaches of scene text detection, stroke width transform was used in \cite{epshtein2010}, stroke feature transform was studied in \cite{huang2013}, MSERs had been used in \cite{Neumann2010} while the well-known FAST corner detector based text fragment detection was reported in \cite{busta2015fastext}. Recent studies of text detection use deep learning based strategies where features are learned automatically with the help convolutional neural networks (CNN). Certain Edge Boxes region proposal algorithm \cite{zitnick2014} and an aggregate channel features detector \cite{dollar2014} had been combined in \cite{jaderberg2016} for detecting words in scene images. This approach made use of the R-CNN object detection framework \cite{girshick2014}. The objective of all these studies was detection of texts in natural scene images where the cameraman controls the camera. However, if the capturing device is a wearable one, detection of incidental texts appearing in such images captured without any control becomes more difficult. In the 2015 version of ICDAR Robust Reading  Competition \cite{karatzas2015icdar}, a challenge on Incidental Scene Text was introduced, based on a dataset of 1,670 sample images captured using the Google Glass. A similar dataset, called COCO-Text \cite{veit2016coco}, was latter introduced in the year 2016. In the next year, Zhou \textit{et al}. \cite{Zhou_2017_CVPR} proposed a fully convolutional network architecture and a non-maximum suppression strategy which can directly locate a word or a text line of arbitrary orientations in similar scene images.  He \textit{et al}. \cite{he2017deep} proposed a direct regression strategy to detect incidental texts of multiple orientations having variations in size and perspective distortions. 
Later Liu \textit{et al}. \cite{Liu_2018_CVPR} combined low-level and high-level feature maps produced by their CNN architecture which could detect incidental texts of different orientations. In \cite{xie2019scene}, a pyramid context network had been proposed for localization of text regions in natural scene images. Wang \textit{et al}. \cite{Wang_2019_CVPR} used a region proposal network together with a RNN for detection of scene text regions of arbitrary shapes. Mask R-CNN has recently been used in \cite{Qin_2019_ICCV} for detection of scene texts of arbitrary shapes. 

\begin{figure*}[!h]
\centering
 \includegraphics[width=6.9in]{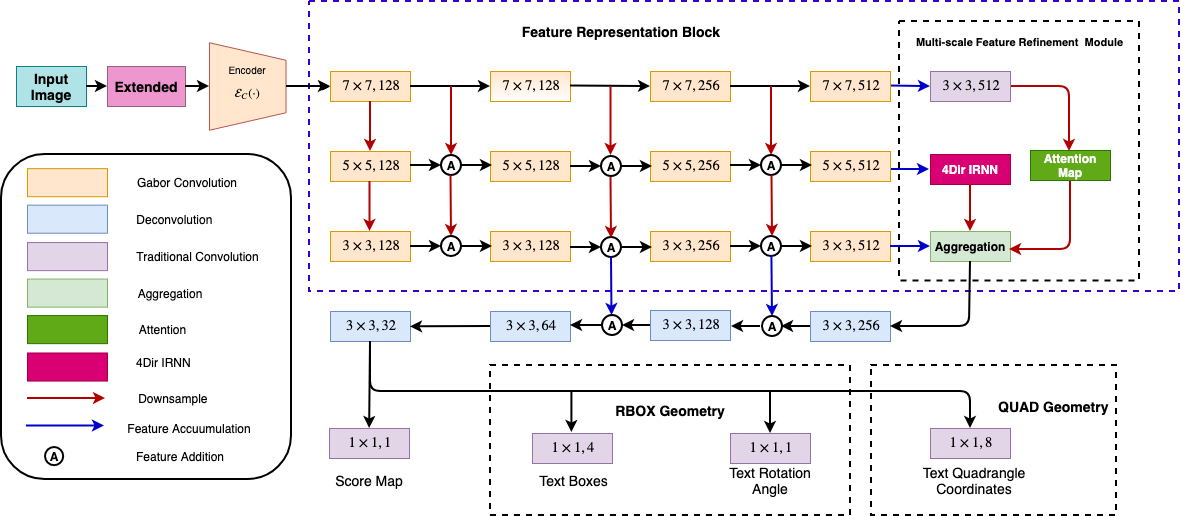}
 \caption{Schematic diagram of the proposed architecture: it contains (i) a ResNet-18 used as feature encoder $\mathcal{E}_{c}(\cdot)$, (ii) a Feature Representation Block (FRB), (iii) a Deconvolution Block and (iv) a Detection Branch. The FRB has a Multi-scale Feature Refinement Module (MFRM) consisting of a 4Dir IRNN, a channel-wise attention block and a CRFs-based aggregation block.} \label{fig:architecture}
 \end{figure*}
\section{Proposed Methodology}
The proposed methodology implements a sophisticated neural network architecture capable of detecting instant scene texts. The network architecture is shown in Figure. \ref{fig:architecture}. It is a fully-convolutional neural network consisting of a block for extension of input image in 4 orientation channels as in \cite{luan2018gabor} followed by a ResNet-18 \cite{He_2016_CVPR} based feature extraction block modulated by Gabor Orientation Filter (GOF), an FRB containing multiple convolutional Gabor Orientation filters and a \textit{Mult-scale Feature Refinement Module} (MFRM), a decoder block of a series of deconvolution operations and finally a detection branch similar to the one used in \cite{Zhou_2017_CVPR}.
\subsection{Feature Representation Block (FRB)}
The \textit{Feature Representation Block} (FRB) receives the output of the {$\mathcal{C}onv_{3\_2}$} layer of backbone Resnet-18 architecture modulated by Gabor Orientation Filter used for enhancement of the robustness of traditional convolution filters towards various image transformations such as scale variations and rotations.
The FRB consists of multiple Gabor Convolutional layers arranged in multiple rows and a \textit{Multi-scale Feature Refinement Module} (MRFM) to make the learned feature representation more abstract. 
Feature maps computed at a higher layer of the FRB is downsampled before entering into the Gabor Convolutional layers of the next row. We also considered kernels of different sizes in each row to tackle the enormous difference in the scales of scene texts. A larger kernel helps to capture more information of Gabor Orientation while a smaller kernel is capable to capture more locally distributed information. 

\subsection{Multi-scale Feature Refinement Module (MFRM)}
MFRM consists of a channel-wise attention map, a 4Dir-IRNN component and an aggregation component modulated by (CRFs). Outputs {\textit($N\times U\times H \times W$)} (where {$N$}, and {$U$} denote respectively the numbers of convolution and orientation channels, {$H$} and {$W$} indicates the height and width of input image) of the last three Gabor Convolutional layers one in each row of FRB are fed as input to the MFRM. We have chosen $N = 512$ and $U = 4$. Note that for every channel $k$, the network produces rank 3 matrix {$\Gamma_{k}$}: {$U \times H \times W$} (where {$k \in [1, 2, \dots 512]$}). We concatenate {$\Gamma_{k}, \forall {k}$} such that the final output matrix will be {$N_{U} \times H \times W$}, where $N_{U} = NU $. Thereafter it has been moved through a {$\mathcal{C}onv_{1\times1}$} reduce the number of channels and fed into the distinct position for every feature. The channel-wise attention module that assists to different channels of the feature map to enhance the essential knowledge, while attenuating noise and background data before transferring it to task-definite layers of the model. We first pass the feature map {$\mathcal{F}_{Conv 3\times3}$} originating from the last convolution block into a {$\mathcal{C}onv_{1\times1}$} to produce a  {$\mathcal{F}_{1\times 1}$} of same size and number of channels. {$\mathcal{F}_{1\times 1}$} is then passed through sigmoidal activation function and multiplied with {$\mathcal{F}_{Conv 3\times3}$} component-wise to produce the feature map {$\mathcal{F}_{att}$}. Mathematically, 
\begin{equation}
    \mathcal{F}_{att}=\mathcal{F}_{Conv 3\times3}\otimes\sigma  (\mathcal{C}onv_{1\times1}(\mathcal{F}_{Conv 3\times3}))
\end{equation}
We follow the 4Dir-IRNN component \cite{Bell_2016_CVPR} in this framework for computing the contextual features. The 4Dir-IRNN component is a sort of RNN composed with ReLU's. The total number of channels in coming from FRB distributed by a factor of 4 and for each 128 channel, we implement a {$\mathcal{C}onv_{1\times1}$} and fed toward the 4Dir-IRNN block. Inside the IRNN we have four RNNs which entirety moves laterally  across the  image in the 4 directions: up, down, left, right. The input-to-hidden is analysed by {$\mathcal{C}onv_{1\times1}$} and shared the convolution layer with 4 recurrent transitions. A distinct direction of RNN will move to collect the contextual features from every 128 channels. The hidden-to-output is merged into a single convolution layer, concatenation followed by a {$\mathcal{C}onv_{1\times1}$} along with a ReLU activation function to generate {$\mathcal{F}_{IRNN}$}. The aggregation block is a multiple features merging unit reinforced with CRFs. Different features transferred from FRB, {$\mathcal{F}_{att}$} and {$\mathcal{F}_{IRNN}$} into the aggregation unit. The CRFs \cite{Liu_2019_ICCV} present in the aggregation component helps to refine each incoming multi-scale feature by combining the corresponding knowledge received from other features. We simply conserve those \textit{($N\times H \times W$)} learned features from the aggregation block and shared into the decoder via {$\mathcal{C}onv_{1\times1}$}.

\subsection{Feature Learning}
Our primary goal is to make a model which is invariant to learn different deformation. Instead of relying on the large diverse dataset, we consider a different approach \textit{Mask and Predict} \cite{kishore2019cluenet} along with pixel-wise blurring strategy for training to learn diverse deformation. The strategy in our proposed network endeavours to explain the difficulty of text detection in cases of occlusion, motion blur and different lighting conditions of the input image. In this approach, we gradually increase a certain percentage of pixel-wise blurring in the input image. Also, we use curriculum learning \cite{bengio2009curriculum} by manually arranging images from simpler to difficult ones. 


\subsection{Feature Decoder}
The feature map containing rich information ejected from the aggregation block is fed into a {$\mathcal{D}econv_{3\times3}$}. In each feature merging point Figure. \ref{fig:architecture}, the incoming feature map from FRB is operated in the same way as described in section 3.2 for its addition with the current feature map. After the two such successive feature additions followed by {$\mathcal{D}econv_{3\times3}$} of 256 and 128 channels, two more {$\mathcal{D}econv_{3\times3}$} of 64 and 32 channels are applied before feeding it to the detection branch.

\begin{figure*}[!h]
 \includegraphics[height=1in]{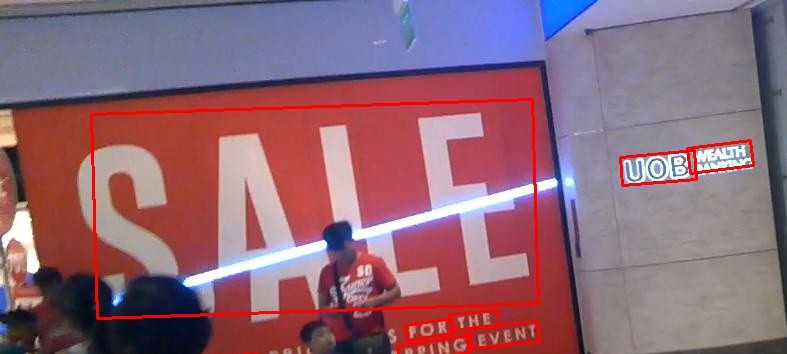}
 \includegraphics[height=1in,width=1.5in]{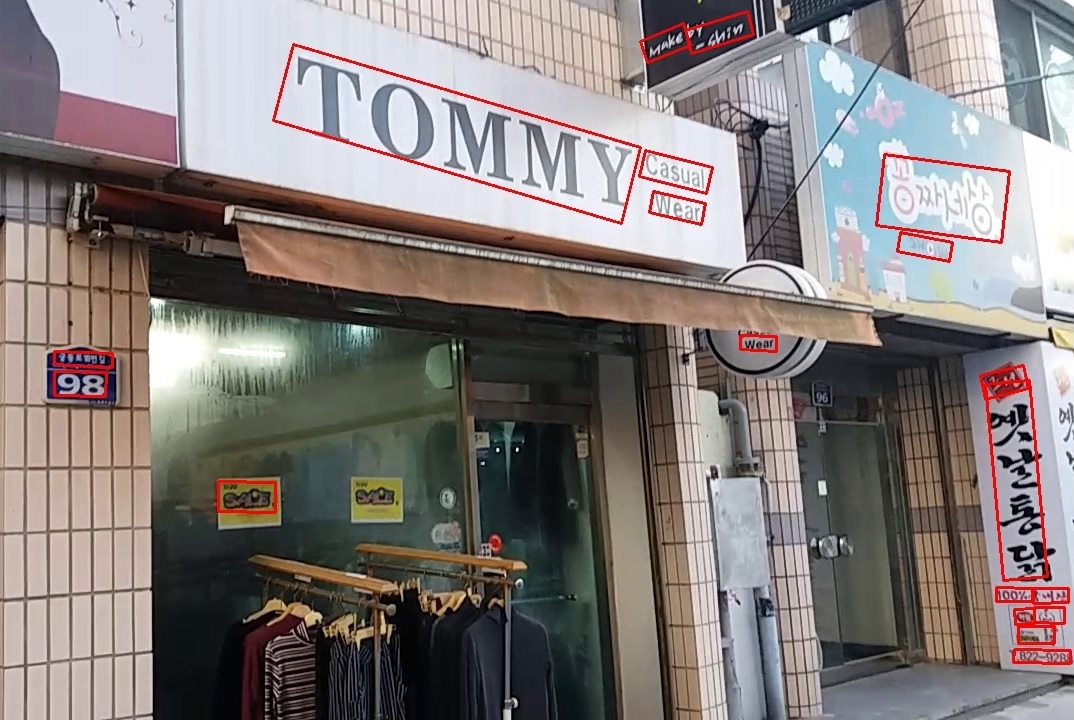}
 \includegraphics[height=1in,width=1.61in]{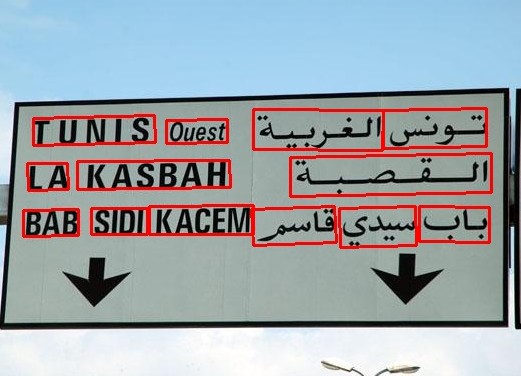}
 \includegraphics[width=1.58in,height=1in]{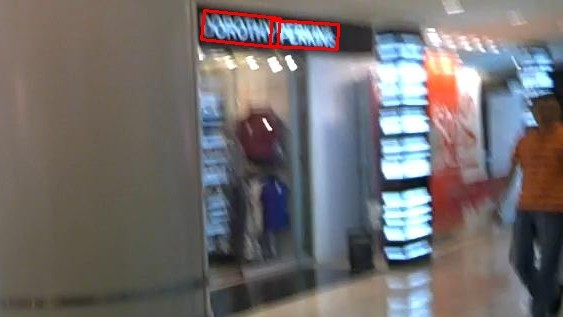}
 \\
 \includegraphics[height=1in,width=2.22in]{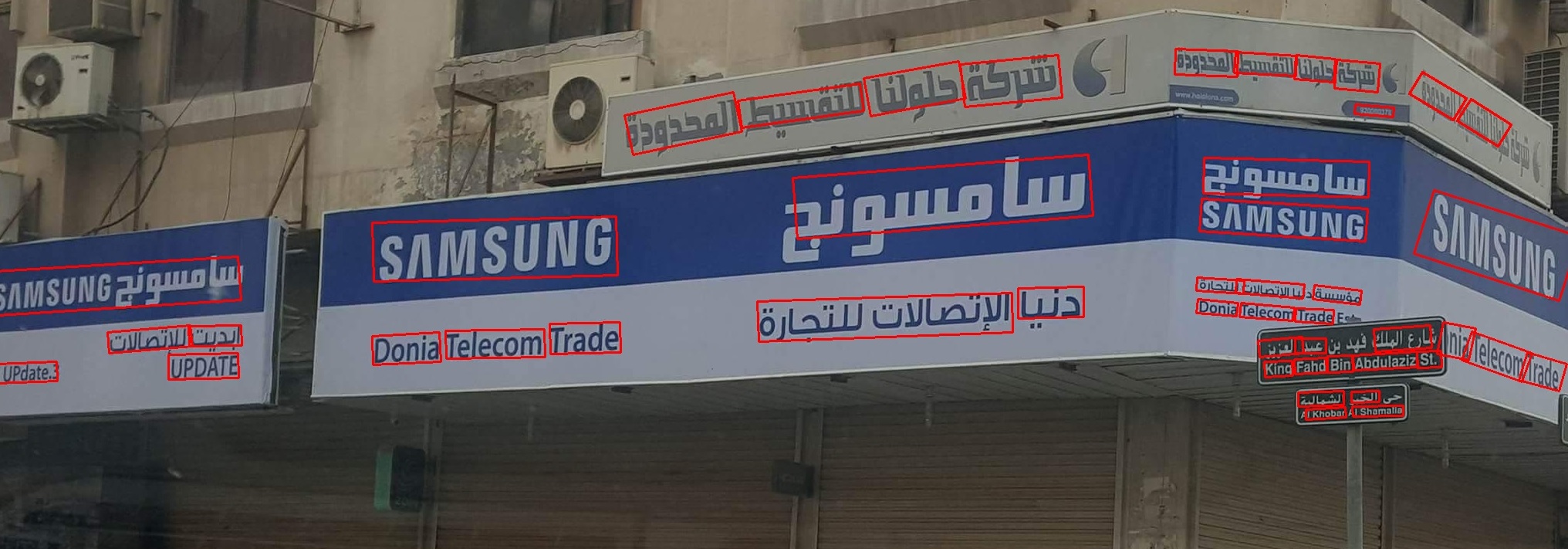}
 \includegraphics[height=1in,width=1.5in]{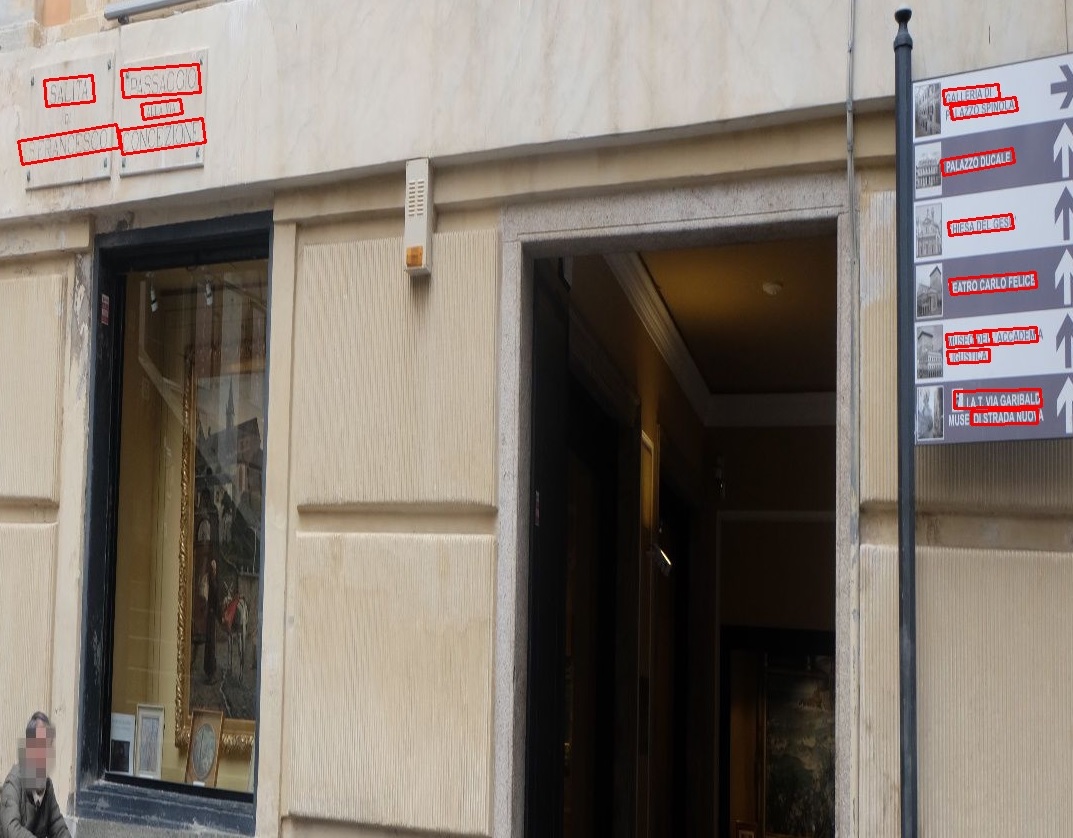}
 \includegraphics[height=1in,width=1.6in]{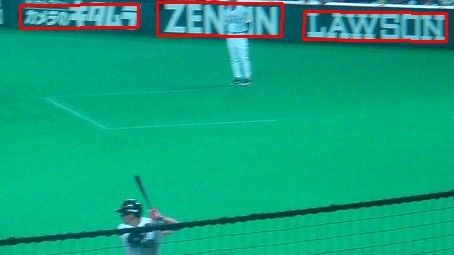}
 \includegraphics[height=1in,width=1.58in]{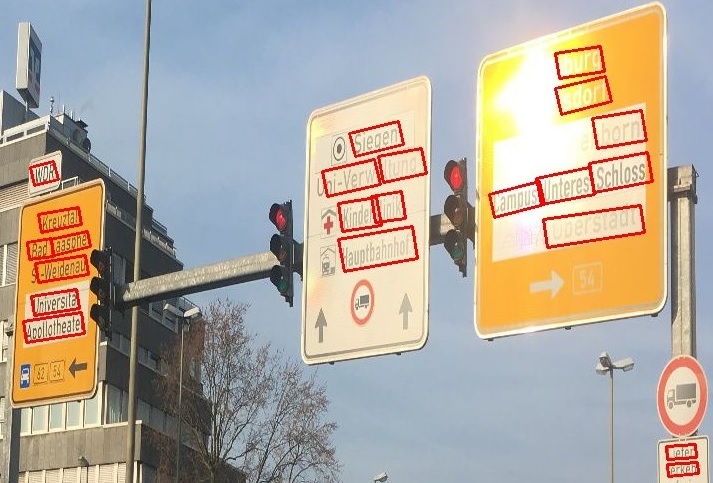}
 \label{fig:3 samples of our detection result}
\caption{Qualitative detection results of the proposed framework on the samples which have texts of widely varying scales, multiple orientations, multiple scripts,  motion blurred,  non-uniformly illuminated,  partially occluded etc.}
\end{figure*}


\subsection{Detection}
The design of the detection branch is similar to the output layer of the network of \cite{Zhou_2017_CVPR}. It consists of several {$\mathcal{C}onv_{1\times1}$} operations of 1, 4, 1, 8 channels of feature maps where the first one provides score map and the following two provide RBOX geometry while the last one provide QUAD geometry. The Loss Function used for the detection branch is as follows,

\begin{equation}
 \mathcal{L} = \mathcal{L}_{s}+\mathcal{\lambda}_{g}\mathcal{L}_{g},
\end{equation}

Where $\mathcal{L}_{s}$ is Loss of the Score Map, $\mathcal{L}_{g}$ is Loss of the Geometry and $\mathcal{\lambda}_{g}$ consider the significance between two losses.  

\begin{table*}[h]
    \centering
    \caption{Comparative detection results of the proposed framework and different SOTA models on various  benchmark sample databases.}
    \label{tab:acc-f1score}
    \resizebox{2\columnwidth}{!}{
    \begin{tabular}{lccccccccccccc}
    \toprule
        Methods & \multicolumn{3}{c}{ICDAR 2015 \cite{karatzas2015icdar}} & \multicolumn{3}{c}{ICDAR 2017 MLT \cite{nayef2017icdar2017} } & \multicolumn{3}{c}{MSRA-TD500 \cite{yao2012detecting}} & \multicolumn{3}{c}{COCO-Text \cite{veit2016coco}} \\
        \cmidrule(r){2-4} \cmidrule(r){5-7} \cmidrule(r){8-10}
        \cmidrule(r){11-13}
         & Recall & Precision & F-Scrore & Recall & Precision & F-Scrore &  Recall & Precision & F-Scrore & Recall & Precision & F-Scrore\\
    \midrule
    EAST \cite{Zhou_2017_CVPR} & 78.3  & 83.3 & 80.7 & - & - & - & 67.4 & 87 .3& 76.0 & 32.4 & 50.3 & 39.4 \\
    He \textit{et al}. \cite{he2017deep} &80.0 & 82.0 & 80.9 & - & - & - & - & - & - &- & - &- \\
    FOTS \cite{Liu_2018_CVPR} &87.9 & \textbf{91.8} & 89.8 & 62.3 & 81.8 & 70.7 & - & - & - & - & - & -\\
    Xie \textit{et al}. \cite{xie2019scene} &85.8 & 88.7 & 87.2 & 68.6 & 80.6 & 74.1 &- & - & - & - & - & -\\
    Wang \textit{et al}. \cite{Wang_2019_CVPR} &86.0 & 89.2 & 87.6 &-& -& -  & \textbf{82.1} & 85.2 & 83.6 &-  &- & - \\
    Qin \textit{et al}. \cite{Qin_2019_ICCV} & 87.9 & 91.6 & 89.7 &- &- &- &- &- &- &- &- &-\\
    SCUT\_ DLVClab \cite{nayef2017icdar2017} &- &- &- & 74.1 & 80.2 & 77.0 &- &- &- &- &- &- \\
    Lyu \textit{et al}. \cite{lyu2018multi} &- &- &- & 70.6 & 74.3 & 72.4 & 76.2 & 87.6 & 81.5 & 32.4 & 61.9 & 42.5 \\
    \textbf{Ours}  &\textbf{89.2} & 91.3 & \textbf{90.2} & \textbf{73.9} & \textbf{88.6} & \textbf{80.5} & 81.6 & \textbf{88.2} & \textbf{84.7} & \textbf{50.6} & \textbf{71.5} & \textbf{59.2} \\
    \bottomrule
    \end{tabular}
    }
    \label{Table 1}
\end{table*}

\section{Experimental Details}
For comparison purpose, the proposed text detection framework has been trained on four publicly available benchmark datasets: ICDAR2015 \cite{karatzas2015icdar}, ICDAR2017 \cite{nayef2017icdar2017}, COCO-Text \cite{veit2016coco} and MSRA-TD500 \cite{yao2012detecting}. Numbers of training samples of these four datasets are respectively 1000, 7200, 43686 and 300 while the volumes of the respective test sets are 500, 9000, 10000 and 200. 

\subsection{Training Details}
 We employ Curriculum Learning \cite{bengio2009curriculum} with respect to progressively increasing pixel-wise blurring, masking percentage and complexity of images to enable the feature learning invariant to image degradation, occlusion and image complexities. We train the Gabor Convolutional layers in the proposed framework with 4 orientation channels and let the scale setting remain 4. Momentum Optimizer is used with an initial learning rate of $0.01$ and momentum of $0.9$ during training. The learning rate is decreased by a factor of 10 after each subsequent 15k iterations. Data Augmentation is also employed to improve the robustness of the proposed framework. The training of the network has been carried out on a computer with two NVIDIA P6 GPU.

\subsection{Evaluation Results}
For performance evaluation of the proposed method, simulations have been done on several benchmark datasets. Among these ICDAR2015 and ICDAR2017 MLT have image samples containing texts affected by motion blur, low-resolution, variable lighting and multi-scaled texts. Simulations done in \cite{Liu_2018_CVPR} ignored image samples suffered from blur during training. However, our model has been purposefully trained on the entire training sets. Results of our extensive simulations have been compiled to construct Table 1.
\section{Conclusion}
In this article, we have presented a novel \textit{Feature Representation Block} (FRB) of a fully convolutional neural network for efficient abstraction of features representing scene texts. For effective learning, we used curriculum learning strategy with respect to percentages of pixel-wise blurring and image sample complexities. Additionally, we used  the \textit{Mask and Predict} strategy to enable the network to produce satisfactory results in cases of partial occlusions. We obtained extensive simulation on four benchmark datasets which include ICDAR 2015, ICDAR 2017 MLT, MSRA-TD500 and COCO-Text. Also, compared these results with the available SOTA performances. The proposed framework can efficiently detect incidental scene texts suffered by uneven lighting condition, blurring, partial occlusions and varying scales. In future, we shall study development of a strategy capable of detecting incidental curved texts. Also, we shall study development of a method to predict occluded texts from the visible partial information.
\bibliographystyle{IEEEbib}
\bibliography{refs}
\end{document}